\documentclass{article}
\usepackage{ijcai23}

\usepackage{times}
\usepackage{soul}
\usepackage{url}
\usepackage[hidelinks]{hyperref}
\usepackage[utf8]{inputenc}
\usepackage[small]{caption}
\usepackage{graphicx}
\usepackage{amsmath}
\usepackage{amsthm}
\usepackage{booktabs}
\usepackage{algorithm}
\usepackage{algorithmic}
\usepackage[switch]{lineno}
\usepackage{amsmath,latexsym,amssymb,amsthm,array,amsfonts,algorithm,booktabs,graphicx,subfigure,multirow,cuted,stfloats}
\usepackage{mathrsfs}
\usepackage{amsfonts,amssymb}
\usepackage{threeparttable}
\usepackage{fancyvrb}
\usepackage{tcolorbox}
\usepackage{listings}
\usepackage{fancyhdr}
\usepackage{lipsum} 

\title{EmoDM: A Diffusion Model for Evolutionary Multi-objective Optimization}

\author{
Xueming Yan$^1$
\and
Yaochu Jin$^{2*}$
\affiliations
$^1$Guangdong University of Foreign Studies\\
$^2$School of Engineering, Westlake University\\
\emails
yanxm@gdufs.edu.cn,
jinyaochu@westlake.edu.cn\\
\vspace{0.8cm}
\textbf{This work has been submitted to the IEEE for possible publication. Copyright may be transferred without notice, after which this version may no longer be accessible.}
}

\begin{document}
\maketitle
\begin{abstract}
Evolutionary algorithms have been successful in solving multi-objective optimization problems (MOPs). However, as a class of population-based search methodology, evolutionary algorithms require a large number of evaluations of the objective functions, preventing them from being applied to a wide range of expensive MOPs. To tackle the above challenge, this work proposes for the first time a diffusion model that can learn to perform evolutionary multi-objective search, called EmoDM. This is achieved by treating the reversed convergence process of evolutionary search as the forward diffusion and learn the noise distributions from previously solved evolutionary optimization tasks. The pre-trained EmoDM can then generate a set of non-dominated solutions for a new MOP by means of its reverse diffusion without further evolutionary search, thereby significantly reducing the required function evaluations. To enhance the scalability of EmoDM, a mutual entropy-based attention mechanism is introduced to capture the decision variables that are most important for the objectives. Experimental results demonstrate the competitiveness of EmoDM in terms of both the search performance and computational efficiency compared with state-of-the-art evolutionary algorithms in solving MOPs having up to 5000 decision variables. The pre-trained EmoDM is shown to generalize well to unseen problems, revealing its strong potential as a general and efficient MOP solver.
\end{abstract}

\section{Introduction}
Numerous real-world optimization problems often involve multiple mutually conflicting objective functions, which are often referred to as multi-objective optimization problems (MOPs) \cite{miettinen1998}. In the past two decades, many evolutionary algorithms for solving MOPs have been proposed \cite{deb_book2001}, which can largely be categorized into decomposition-based \cite{zhang2007moea,cheng2016reference}, Pareto-based \cite{deb2002fast}, and performance indicator-based \cite{beume2007sms}. Multi-objective evolutionary algorithms (MOEAs) have been shown successful in solving a wide range of real-world MOPs, including data mining \cite{barba2018jmetalsp}, decision-making \cite{fioriti2020economic}, engineering design \cite{terra2018stratification}, and machine learning \cite{Dushatskiy2023icml,jin_book2006}. Despite the great promise they offer, however, most MOEAs commonly rely on a large number of evaluations of the objective functions, making it prohibitive to use them for solving expensive real-world problems. 

Much effort has been dedicated to reducing function evaluations required by MOEAs. One early line of research is surrogate-assisted evolutionary algorithms \cite{jin2011survey,li2022survey}, where machine learning models are adopted to build surrogates to estimate the objectives. More recently, deep generative models such as generative adversarial networks and transformers have been employed to either assist the evolutionary search \cite{he2021gan,hong2024transformer,liu2023llm} or replace the evolutionary optimizer \cite{liu2023large}. In \cite{lin2022iclr}, an attention model is adopted to learn the relationship between an arbitrary preference and its corresponding Pareto solution for combinatorial optimization. As result, a new Pareto solution can be approximated without further search for the same multi-objective optimization problem when the preference is changed. As a step further, a data-driven end-to-end approach to multi-objective combinatorial optimization using deep graph neural networks has been reported in \cite{liu2023end}. This approach distinguishes itself with most machine learning-assisted evolutionary algorithms in that it trains graph neural networks on a set of previously solved instances of the same problem, e.g., the travelling salesman problem (TSP) with different numbers of cities and locations, and then directly solves new instances without evolutionary search. This way, the required function evaluations can be significantly reduced because only one set of candidate solutions need to be evaluated. However, this approach may suffer from poor generalization across different instances and usually additional local search is needed \cite{liu2023end}.

Recently, generative diffusion models (DMs) demonstrated remarkable performance across multiple applications, including image synthesis \cite{dhariwal2021diffusion}, molecular graph modeling \cite{huang2023mdm}, and material design \cite{karras2022elucidating}. These DMs excel at modelling complex datasets and generating high-quality samples for performing diverse tasks. So far, not much work on using DMs for optimization has been reported with a few exceptions. For example, Graikos {\textit et al.} \cite{graikos2022} proposed to solve the TSP using an image-based DM by considering solutions of the TSP as grey-scale images. Sun and Yang \cite{sun2023difusco} went a step further by employing a graph DM for combinatorial optimization.  

\begin{figure*}[!t]
\centering
\includegraphics[width=5.9in]{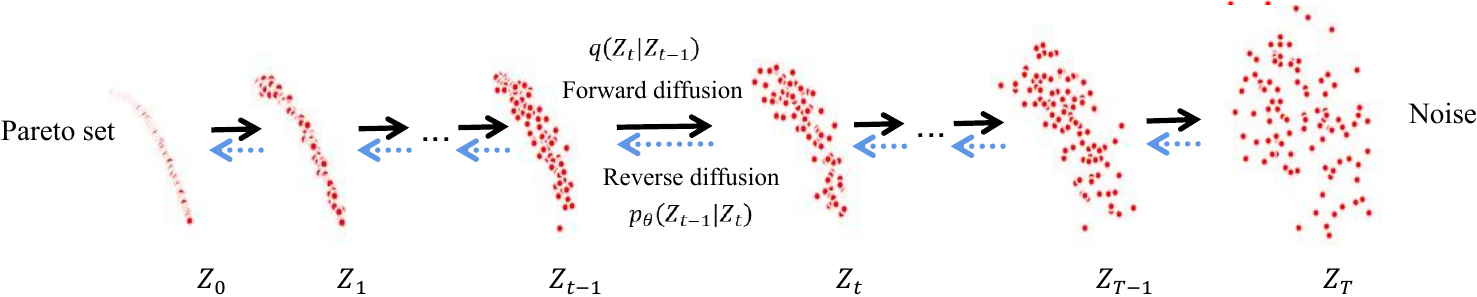}
\caption{An example of the forward and reverse diffusion processes of a DM for evolutionary multi-objective optimization.}
\label{fig1}
\end{figure*}

Different from the above ideas, this work proposes a DM for evolutionary multi-objective optimization without evolution, called EmoDM. The main idea is to learn the evolutionary search capability with a DM from previous evolutionary search processes. Once it has been trained on a range of MOPs, EmoDM can approximate a set of Pareto optimal solutions for any new MOP starting from randomly generated solutions without additional evolutionary search, thereby significantly reducing the objective function evaluations. An illustrative example of the forward and reverse diffusion of the EmoDM is given in Fig. \ref{fig1}, where $Z_t = \{\mathbf{x}^{0}(t),..., \mathbf{x}^{j}(t),..., \mathbf{x}^{N}(t)\}$ represents a population of solutions at step $t$, $\mathbf{x}^{j}(t)$ denotes the $j$-th solution in the population ($j=1,..., N$), and $N$ is the population size. The forward diffusion is the reversed evolutionary search process, i.e., $Z_T$ is the initial population in the evolutionary search, which is typically randomly generated, and $Z_{T-1}$ is the second, and $Z_{0}$ is the final generation. In training EmoDM, we use solutions from two consecutive generations of the evolutionary search, e.g., $Z_{t}$ and $Z_{t-1}$), to estimate the noise distributions. By contrast, the reverse diffusion can be seen as the process of the forward evolutionary search, which will result in a set of Pareto optimal solutions starting from a randomly generated population. 

The contributions of this paper include:
\begin{itemize}
    \item We propose a pre-trained diffusion model, EmoDM, to solve multi-objective optimization problems. The key idea is use the solutions in every two consecutive generations of the evolutionary search process in an reversed order to learn the noise distribution of each forward diffusion step for each objective. The trained EmoDM can then generate a set of approximate Pareto solutions for new MOPs with reverse diffusion. 

    \item To enhance the scalability of EmoDM to the search dimension, we propose an mutual entropy-based attention mechanism for capturing the most important decision variables in sampling. As a result, EmoDM performs well also on high-dimensional MOPs.  

    \item EmoDM is examined on two suites of widely used benchmark problems, demonstrating its competitive performance on MOPs with up to 5000 decision variables in comparison to four representative MOEAs. 
\end{itemize}

\section{ Preliminaries}
\subsection{Multi-objective Optimization}
Without loss of generality, the unconstrained MOP can be expressed as follows:
 \begin{align} 
 \label{obj1}
 &  \mathop{\text{min}}\limits_{\mathbf{x} \in \Omega } F(\mathbf{x}) = (f_{1}(\mathbf{x}),f_{2}(\mathbf{x}),...,f_{m}(\mathbf{x}) )^{T}, 
 \end{align}
 where $\mathbf{x} = (x_{1},x_{2},...,x_{d})$ denotes $d$ decision variables in the search space $\Omega  \subset \mathbb{R}^{d}$, $m$ represents the number of objectives, and $f_{i}(\mathbf{x})$ is the $i$-th ($i=1,2,...,m$) objective function. In non-trivial scenarios, the decision variables frequently exhibit interdependencies, and the objectives commonly manifest conflicts. Specifically, there is no single solution capable of concurrently minimizing all objectives. Therefore, the goal of optimizing an MOP is to obtain a set of Pareto optimal solutions within its search space \cite{miettinen1998}. 

For two solutions $\mathbf{x}_{1}, \mathbf{x}_{2} \in \Omega$, $\mathbf{x}_{1}$ is said to dominate $\mathbf{x}_{2}$, denoted as $\mathbf{x}_{1} \prec \mathbf{x}_{2}$, if and only if $f_{i}(\mathbf{x}_{1}) \le  f_{i}(\mathbf{x}_{2})$, for all $i \in  \{1, 2,..., m\} $, and there exists at least one $j \in \{1, 2,..., m\}$ such that $f_{j}(\mathbf{x}_{1}) < f_{j}(\mathbf{x}_{2})$. $\mathbf{x}_{1}$ is a Pareto optimal solution if and only if there does not exist $\forall \mathbf{x}_{2} \in \Omega $ that dominates it. The entire collection of Pareto solutions in the objective space is called Pareto front (PF). In multi-objective optimization, we aim to closely approximate and evenly cover the PF of the MOP.

\subsection{Diffusion Models}
Diffusion models (DM) can be seen as a type of variational inference \cite{NEURIPS2021_b578f2a5}, which employs a denoising network to progressively converge towards an approximation of a real sample $z \sim q(z)$ across a sequence of estimation steps. Here, we abuse the notation $z$ and use it to denote an element of data, $q(z)$ is the distribution \cite{ho2020denoising}. DM can be divided into a forward and a reverse diffusion process to be elaborated below.
\subsubsection{Forward diffusion process}
The forward diffusion process is defined by a Markov chain consisting of $T$ steps, from the original data $z_{0}$ to the final $z_{T}$. At the $t$-th step, Gaussian noise with a variance of $\beta_{t}$ is introduced to $z_{t-1}$, producing $z_{t}$ with the distribution $q(z_{t}|z_{t-1})$, which is expressed by
\begin{align} 
 &  q(z_{t}|z_{t-1}) = \mathcal{N} (z_{t};\mu_{t},\Sigma_{t}),
 \end{align}
where $q(z_{t}|z_{t-1})$ is a normal distribution characterized by the mean $\mu_{t}$ and the variance $\Sigma_{t} = \beta_{t}I$. Here, $I$ is the identity matrix, signifying that each dimension has the same standard deviation $\beta_{t}$. 
The posterior probability can be expressed in a tractable form as
\begin{align} 
 \label{obj2}
 &  q(z_{1:T}|z_{0}) = \prod_{t=1}^{T}  q(z_{t}|z_{t-1}),
 \end{align}
Then, the transition from \(z_{0}\) (original data) to \(z_{t}\) (the $t$-th step noisy data) can be expressed as
\begin{equation}
 z_{t} = \sqrt{1-\beta_{t}} z_{t-1} + \sqrt{\beta_{t}} \epsilon_{t-1} = \sqrt{\alpha _{t}} \\ z_{t-2} + \sqrt{1 - \alpha_{t}} \epsilon_{t-2}  \\
 = ... = \sqrt{\bar{\alpha }_{t}} z_{0} + \sqrt{1 - \bar{\alpha }_{t}} \epsilon_{0}, \quad\quad\quad\quad\quad\quad\quad\quad\quad\quad\quad
\end{equation}
where \(\epsilon_{0}, ...,\epsilon_{t-1} \sim \mathcal{N}(0,I)\) is the noise added at each step. $\bar{\alpha } = \prod_{j=1}^{t} \alpha_{j}$  and $\alpha_{t} = 1 - \beta_{t}$ is a variance schedule. 
Consequently, $z_t$ can be represented using the following distribution.
\begin{equation}
 z_{t} \sim q(z_{t}|z_{0}) = \mathcal{N}(z_{t};\sqrt{\bar{\alpha }_{t}} z_{0}, (1 - \bar{\alpha }_{t})I ).
\end{equation}
Given $\beta_t$ as a hyperparameter, precomputing $\alpha_t$ and $\bar{\alpha}_t$ for all timesteps allows noise sampling at any $t$ to obtain $z_t$. The variance parameter $\beta_t$ can be fixed or chosen via a $\beta_t$-schedule over $T$ timesteps \cite{ho2020denoising}.

\subsubsection{Reverse diffusion process}
The reverse diffusion is regarded as the denoising process, aiming to learn the reverse distribution $q(z_{t-1}|z_{t})$ with a parameterized model $p_{\theta }$:
\begin{equation}
 \label{obj3}
 p_{\theta }(q(z_{t-1}|z_{t})) = \mathcal{N}(z_{t-1}; \mu_{\theta}(z_{t},t), \Sigma_{\theta}(z_{t},t) ).
\end{equation}
When incorporating the conditional information, $p_{\theta }(q(z_{t-1}|z_{t}))$ can be represented as a noise prediction model with a fixed covariance matrix as 
\begin{equation}
  \Sigma_{\theta}(z_{t},t) = \beta_{t}I,
\end{equation}
and the mean is formulated as
\begin{equation}
\mu_{\theta}(z_{t},t) = \frac{1}{\sqrt{\alpha_t}} \left( z_t - \frac{\beta_t}{\sqrt{1-\bar{\alpha_{t}}}} \epsilon_\theta(z_t, t) \right).    
\end{equation}
The reverse diffusion chain is parameterized by
$\theta$ as 
\begin{equation}
z_{t-1}|z_{t} = \frac{1}{\sqrt{\alpha_t}} \left( z_t - \frac{\beta_t}{\sqrt{1-\bar{\alpha_{t}}}} \epsilon_\theta(z_t, t)\right) + \sqrt{\beta_{t}}\epsilon,
\label{eq9}
\end{equation}
where the noise $\epsilon \sim \mathcal{N}(0,I)$, and $t\in [1,T]$. Subsequently, we can obtain the trajectory from $z_{T}$ to $z_{0}$ as 
\begin{equation}
p_{\theta}(z_{0:T}) = p_{\theta}(z_{T})\prod_{t=1}^{T}p_{\theta }(z_{t-1}|z_{t}). 
\end{equation}
Once we learn the reverse distribution $q(z_{t-1}|z_{t})$, we can sample $z_{T}$ from $\mathcal{N}(0,I)$, execute the reverse process, and obtain a sample from $q(z_{0})$. 

 
\section{The Proposed Method}
\subsection{Overall framework}
\begin{figure}[t]
    \centering
    \includegraphics[width=\linewidth]{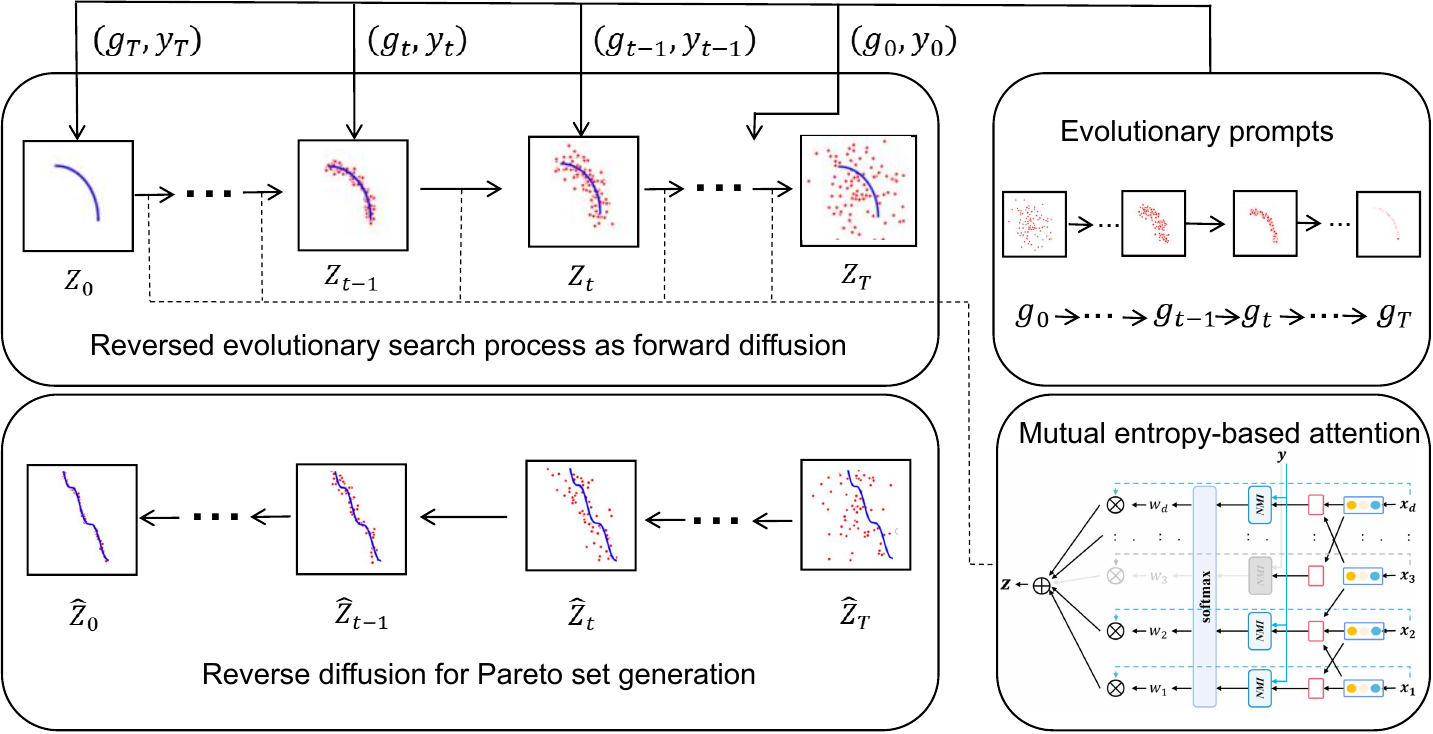}
    \caption{A framework of the proposed EmoDM, where the reversed evolutionary search process (from the final generation to the first generation) is used as the training data (evolutionary prompts) to learn a sequence of Gaussian noise models. An mutual entropy-based attention mechanism is introduced for focusing on the most important decision variables. After training, EmoDM can approximate a set Pareto solutions for a new MOP by gradually reducing the noise based on the learned noise distributions.
 }
    \label{fig2}
\end{figure}

The overall framework of the proposed EmoDM for multi-objective optimization is depicted in Fig. \ref{fig2}. Similar to DDPM \cite{ho2020denoising}, EmoDM consists of two stages, a forward diffusion that learns the reverse evolutionary search from the converged final population to the randomly initialized population  and a reverse diffusion that approximates a Pareto set starting from a randomly generated initial population.  

As explained in \cite{igel2007covariance}, the Markov chain can be employed to model the evolutionary search dynamics in multi-objective optimization. Thus, given the input data ${Z}_{0}$, which is a set of Pareto solutions of an MOP found by an MOEA, the evolutionary prompts, i.e., solutions of two consecutive populations, ($g_{t}, \mathbf{y}_{t}$) can be utilized to estimate the noise data step by step, where $g_{t}$ stands for the population data of $(T-t)$-th generation used to generate sample ${Z}_{t}$ at $t$-th step. $\mathbf{y}_{t} = (f_1({g}_{t}),f_2({g}_{t}),...f_m({g}_{t}))$ represents the $m$ objective functions (vectors) of the current population. The total number of steps ($T$) in EmoDM equals to the total number of generations of evolutionary optimization. Integrating the evolutionary prompts with an mutual entropy-based attention mechanism, which extracts important latent search dimensions, we can estimate the multivariate noise distribution between two consecutive steps by training the forward diffusion models on each given sample. With the guidance of the difference in mutual information entropy between decision variables and objectives, we can reconstruct the Pareto set for a new MOP in the reverse diffusion, where sampling is achieved by iteratively reducing the influence of multivariate noise.


\subsection{Reversed evolutionary search process as the forward diffusion}

The population dynamics of an evolutionary algorithm, i.e., the distributions of the individuals (candidate solutions) in the search space, driven by the genetic variations, such as crossover and mutation, and environmental selection, can be seen as a denoising process from a randomly generated population. Here, we have two basic assumptions. First, the changes of the population dynamics from one generation to the next (in the reversed order) in solving a given MOP, which essentially represents the evolutionary search behavior, can be learned by the EmoDM as the forward diffusion. Thus, the trained EmoDM can replicate the evolutionary search process of the given MOP with reverse diffusion. Second, if the EmoDM is trained on multiple MOPs, it is able to generalize to solve new MOPs with the learned search behaviors in a computationally more efficient way compared with an EMOA. 

Algorithm \ref{alg1:algorithm} delineates the forward training process based on the reversed evolutionary search process. It begins with the design of the evolutionary prompts (the data of every two generations, refer to Fig. \ref{fig1}) for training the EmoDM. In addition, the mutual entropy-based attention mechanism is utilized to explore the mutual information between the decision variables and the objectives to reduce the search space. In the forward diffusion stage, we train a series of noise distributions for each multi-objective optimization (MOP) problem.

\begin{algorithm}[H]\small
    \caption{Forward training process based on evolutionary prompts}
    \label{alg1:algorithm}
     \textbf{Input}: a sequence of evolutionary prompts $(g_{i}, y_{i})$, $i = 1,2,...,K$, where $K$ is the number of MOPs to be trained \\
     \textbf{Output}: the pre-trained EmoDM
    \begin{algorithmic}[1] 
        \STATE Initialize the training data with evolutionary prompts  $(g_{i}, \mathbf{y}_{i})$.
        \STATE Apply the mutual entropy-based attention to capture the important decision variables according to Eq. (\ref{lab:mut}). 
        \STATE \textbf{for}  {$1 \le i \le K$} \\
        \STATE  \quad Estimate the Gaussian noise across steps repeat until the model of the noise distribution converges. \\
        \STATE \textbf{end~for} 
    \end{algorithmic}
\end{algorithm}

\subsubsection{Evolutionary prompts as training data}

Given an MOP, which has $m$ objectives and $d$ decision variables. The population size is $N$ and the maximum number of generations is $T$, which is the number of steps of diffusion model. In this study, two types of information are provided in our prompt at the $t$-th step of the forward diffusion, which include the decision variables of $N$ individuals (solutions) ${g}_{t}$ at generation $t$ from the reversed evolutionary process, and the $m$-dimensional objective vectors $\mathbf{y}_{t}$ related to the $N$ individuals. 

\subsubsection{Mutual entropy-based attention}
In this section, we employ a mutual entropy-based attention mechanism to capture the importance of the decision variables on the objectives from evolutionary prompts, which aims to focus on the decision variables having a significant average impact on all objectives.
Let $Z_t = \{\mathbf{x}^{0}(t),..., \mathbf{x}^{j}(t),..., \mathbf{x}^{N}(t)\}$ represent the sample with $N$ data points $\mathbf{x}^{j}(t)$ at $t$-th step. Data points $\mathbf{x}^{j}(t)$ is regarded as observations of the decision variable $\mathbf{x}_t$ of the MOP problem. 
The mutual information entropy, denoted as $\mathit{I} (\mathbf{x}_{t} ;\mathbf{y}_{t})$, is defined as follows:
\begin{equation}
  \mathit{I} (\mathbf{x}_{t} ;\mathbf{y}_{t}) = \sum_{\mathbf{x}_{t}}\sum_{\mathbf{y}_{t}} p(\mathbf{x}_{t}, \mathbf{y}_{t})log (\frac{p(\mathbf{x}_{t},\mathbf{y}_{t})}{p(\mathbf{x}_{t}).p(\mathbf{y}_{t})}),
\end{equation}
where $\mathbf{y}_{t} = [f_{1}(\mathbf{x}_{t}), f_2(\mathbf{x}_{t}), ..., f_m(\mathbf{x}_{t})]$ are the objectives, $p(\mathbf{x}_{t})$ and $p(\mathbf{y}_{t})$ represent the marginal probability density of the decision variable and objective variables, respectively. $p(\mathbf{x}_{t},\mathbf{y}_{t})$ stands for the joint probability density. In this study, $p(\mathbf{x}_{t})$, $p(\mathbf{y}_{t})$ and $p(\mathbf{x}_{t},\mathbf{y}_{t})$ can be estimated by employing a statistical method \cite{ren2023multi} using evolutionary prompts.
If the mutual information entropy is zero, it indicates that the observations of the decision variables do not provide any additional information about the objectives. A larger mutual information entropy may suggest that the influence of decision variables on the objective functions is more complex or intimate. The average normalized mutual information entropy between each decision variable $\mathbf{x}_{t}(i)$ and the objective vector $\mathbf{y}_{t}(j)$ is calculated as follow:
\begin{equation}
 AvgNMI(\mathbf{x}(i)) = \frac{1}{m} \sum_{j=1}^{m} \mathit{I} (\mathbf{x}_{t}(i) ; \mathbf{y}_{t}(j)).
\end{equation}
Thus, the soft-attention coefficient vector $\mathbf{w_{t}}$ at current step is calculated as follows to focus on the most important decision variables:
\begin{equation}
  \mathbf{w}_{t} = \text{softmax} (AvgNMI(\mathbf{x}(i))).
\end{equation}
Finally, we take the weighted sum of the decision variables using soft attention weights $\mathbf{w}_{t}$ and the observed data point $\mathbf{x}^{j}(t)$ in sample $Z_{t}$ can then be recomputed by
\begin{equation}
  \mathbf{x}^{j}(t) = \mathbf{w}_{t} \mathbf{x}^{j}(t).
\label{lab:mut}
\end{equation}
As a result, the mutual entropy-based attention can help EmoDM focus on the crucial decision variables by adjusting the weights of different decision variables.

\subsubsection{Noise estimation}
The objective of forward training process is to minimize the Mean Squared Error (MSE) loss defined by
\begin{equation}
\label{eq151}
L = \frac{1}{T} \sum_{i=1}^{T}\left \| \epsilon_t -\epsilon_\theta({Z}_t, t)  \right \| ^{2}. 
\end{equation}
Let $\epsilon_t$ be the Gaussian noise added at each step,  then the estimated noise $\epsilon_\theta(Z_t, t)$ can be obtained by the back-propagation algorithm \cite{lecun1989backpropagation}. The transition from $Z_{t-1}$ to $Z_{t}$ can be expressed as 
\begin{equation}
 {Z}_{t} = \sqrt{\alpha_{t}} {Z}_{t-1} + \sqrt{1 - \alpha_{t}} \epsilon_{t},
\end{equation}
where ${Z}_{t}$ and ${Z}_{t-1}$ can be obtained by  Eq. (\ref{lab:mut}), and $\alpha_{t}$ is the variance schedule of the noise $\epsilon_{t}$. Consider the distinctiveness of the decision variables in the samples, we employ the average variance schedule of different decision variables as the variance schedule  $\alpha_t$ of the multivariate Gaussian distribution noise $\epsilon_t$ at the $t$-th step. 


\subsection{Pareto set generation}
Through the forward training, we have learned a sequence of noise models based on the evolutionary prompts collected from solving MOPs. Next, we will generate a Pareto set for a new MOP in the reverse diffusion process. In reverse diffusion, we employ a sampling strategy based on pre-trained EmoDM together with the attention mechanism to gradually eliminate noise, thereby restoring the Pareto sets for the new problem. Algorithm \ref{alg2:algorithm} shows the process of a mutual entropy-based sampling. First, we randomly generate initial noise data $\hat{{Z}}_{T} \sim \mathcal{N}(0,I)$ at the $T$-th step. Based on the pre-trained EmoDM, the normalized mutual information entropy between the decisive variables and objective objectives is considered to sample the Gaussian noise for the corresponding steps. 
In the sampling process, we simulate the denoising process of the training data between each step by finding out the previously learned noise that is most similar to the current one in terms of mutual information entropy:
\begin{equation}
\label{eq16}
\min_{i=1}^{K} | NMI(\mathbf{\hat{x}}_{t} ;\mathbf{\hat{y}}_{t}) - NMI(\mathbf{x_{t}}(i) ;\mathbf{y_{t}}(i)|,
\end{equation}
where $\mathbf{\hat{x}}_{t} $ and $\mathbf{\hat{y}}_{t}$ represents the decision variables and multi-objective vectors at the $t$-th step, respectively,
$\mathbf{x_{t}}(i)$ and $\mathbf{y_{t}}(i)$ stands for the decision variables and multi-objective vectors corresponding to the $t$-th step in the $i$-th evolutionary prompts, $K$ denotes the number of MOPs used in forward training. It should be pointed out that a similar problem will be re-identified only when $t=2^i$-th step, where $i=0,1, 2, ...,\left \lfloor  log_{2}T  \right \rfloor$. That is, the last step to re-identify a similar problem is in the first step. This way, we can generate a sequence of candidate solutions and a set of Pareto solutions can be approximated through  $T$ steps of the reverse diffusion.

\begin{algorithm}[tb]\small
    \caption{Mutual entropy-based sampling for solution generation}
    \label{alg2:algorithm}
    \textbf{Input}: a MOP $p_{new}$ \\
    \textbf{Output}: a generated Pareto set ${\hat{Z}}_{0}$ 
    \begin{algorithmic}[1] 
        \STATE Generate initial noisy data ${\hat{Z}}_{T} \sim \mathcal{N}(0,I)$.
        \STATE Calculate the normalized mutual information entropy for ${\hat{Z}}_{T}$.
        \STATE $t = T, s = 2^{\left \lfloor  log_{2}{T}  \right \rfloor}$.
        \WHILE{$t > 0$}
        \IF{$t = T || t=s$}
        \STATE Select the most similar prompt having the smallest difference in mutual information entropy by Eq. (\ref{eq16}).
        \STATE Employ the forward Gaussian noise at the $t$-th step from the pre-trained EmoDM. 
        \STATE s = s/2.
        \ELSE
         \STATE Sample the forward Gaussian noise based on the current evolutionary prompt at $t$-th step from the pre-trained EmoDM.
        \ENDIF
        \STATE Generate ${\hat{Z}}_{t-1}$ according to Eq. (\ref{eq9}).
        \STATE t = t+1.
        \ENDWHILE
    \end{algorithmic}
\end{algorithm}

Note that in the steps of reverse diffusion when Equation (\ref{eq16}) is used to identify the trained problem that is most similar to the current problem, function evaluations are needed. Thus, if we perform this calculation in every step, the same number of function evaluations as the MOEAs will be required. In the main experimental studies, we empirically set the frequency (denoted by $\xi$) at which EmoDM needs to check the problem similarity to be in every $t=2^i$-th steps, where $i$ is the step index. We will further analyze the influence of hyperparameter $\xi$ on the performance of EmoDM in Section \ref{sec:further}.  
\section{Experiments}
\subsection{Experimental settings}
To evaluate the effectiveness of the proposed EmoDM, we conducted experiments on one MOP benchmark, the WFG test suite \cite{hubandscalable}, and one large-scale MOP test suite, LSMOP \cite{cheng2016test}. Three representative MOEAs, namely NSGA-II \cite{deb2002fast}, MOEA/D \cite{zhang2007moea}, and SMS-EOMA \cite{beume2007sms}, are employed for comparison with the proposed EmoDM. A widely used performance indicators, called the inverted generational distance (IGD), is adopted to evaluate the performance of the compared algorithms. For fair comparisons, we utilize the recommended parameter settings for the compared algorithms as reported in the literature. 

In this study, we utilize evolutionary prompts as the training data for our EmoDM. NSGA-II is employed to optimize MOPs on a software tool PlatEMO \cite{tian2017platemo} and the solutions in all generations are recorded so that they can be used for training EmoDM. To investigate the impact of the number of objective functions, we consider both two- and three-objectives cases. In the first set of experiments, EmoDM is trained on six 30-dimensional (30-D) ZDT instances and seven 30-D DTLZ instances \cite{deb2005scalable}, and then tested on 30-D WFG instances. By contrast, EmoDM is trained on 200-D LSMOP1, LSMOP2, LSMOP3 and LSMOP4, LSMOP9 instances and then tested on 200-D LSMOP5, LSMOP6, LSMOP7 and LSMOP8 instances. In the experiments, the decision variable dimensions used in the training problems are set to be consistent with those of the testing MOPs. In evolutionary optimization, the population size $N$ is set to 100, and the maximum number of generations $T$ is set to 200 for the WFG instances and 2000 for LSMOP instances, which is also the number of steps in forward diffusion of EmoDM. 

In the second set of experiments, we validate the effectiveness of EmoDM on large-scale problems, where the dimension of the search space is set to 500, 1000, 3000, and 5000, respectively, for LSMOP instances in both training and test.


\subsection{Overall performance}
Tables \ref{tab:1} presents the average IGD values for the 2-objective WFG instances and 3-objective four LSMOP instances. 
To be consistent with the setting of the MOEAs, we also conduct 30 independent runs for EmoDM to generate Pareto sets on each MOP instance. 
In the tables, the best-performing result on each instance is highlighted in bold. The symbols ``$+$,'' ``$-$,'' and ``$=$'' indicate that the performance of the corresponding approaches, in terms of IGD, is better than, worse than, or comparable to that of EmoDM in solving the MOP instances.

EmoDM demonstrates highly promising performance in terms of the average value of IGD when compared to NSGA-II, MOEA/D, and SMS-EOMA, three popular and representative MOEAs, as shown in Table \ref{tab:1}. Out of a total of 13 test instances, EmoDM achieves the best IGD results on 10 instances, followed by NSGA-II with 2 best results and SMS-EOMA with 1 best result. EmoDM significantly outperforms NSGA-II, MOEA/D, and SMS-EOMA on 11, 12, and 12 test instances, respectively. Notably, EmoDM performs the best on all LSMOP test instances in terms of IGD. These results are quite surprising, since the search capability of EmoDM is learned from NSGA-II. This might be attributed to the fact that EmoDM has been trained on multiple instances and implicit knowledge transfer helps improve its search performance on new instances. 

\begin{figure*}
  \centering
   \includegraphics[width=0.8 \linewidth]{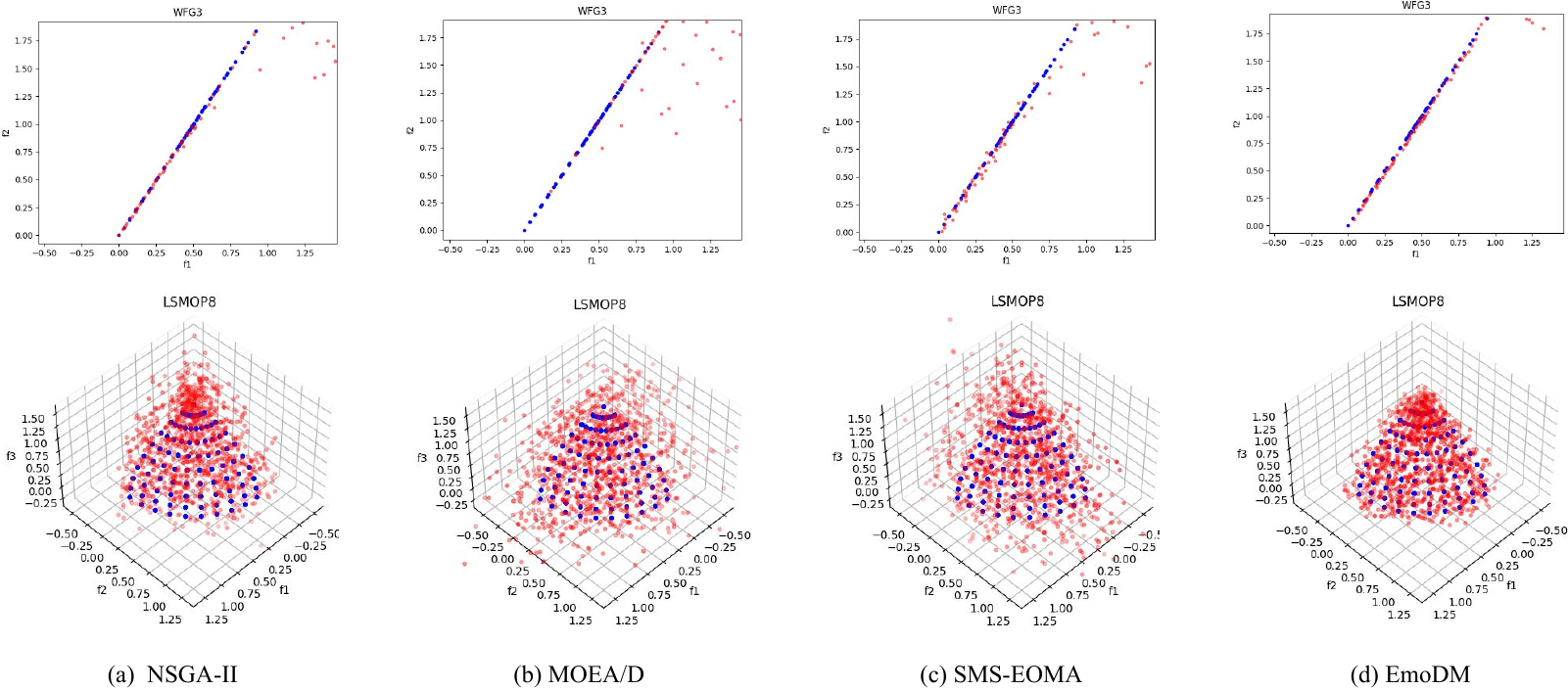}
   \caption{Approximated PFs on the 2-objective WFG3 instance and 3-objective LSMOP8 instance.}
   \label{fig:pf3}
 \end{figure*}

We also visualize the approximated Pareto front obtained by the four algorithms under comparison on the 2-objective WFG3 and 3-objective LSMOP8 in Fig. \ref{fig:pf3}, where the best results from 30 runs are plotted. These results confirm the convergence performance of EmoDM. 

\begin{table}[]\tiny 
\centering

\setlength{\tabcolsep}{4pt}
\begin{tabular}{cllll} \toprule
\multicolumn{1}{l}{Problem} & NSGA-II & MOEA/D & SMS-EMOA & EmoDM  \\ \hline
WFG1  & 6.88e-1(7.32e-3)-	& 1.03e+0(1.45e-3)-	& 6.56e-1(7.84e-3)-	& \textbf{6.55e-1(3.25e-3) }     \\ 
WFG2 & 6.76e-2(8.14e-3)-    & 2.05e-1(3.21e-3)-	& 5.05e-2(6.73e-3)-	&\textbf{4.14e-2(2.17e-3) }      \\ 
WFG3 & 7.56e-2(8.94e-3)-	    & 2.09e-1(2.68e-3)-	& 4.85e-2(5.79e-3)-	& \textbf{4.56e-2(3.64e-3)}       \\ 
WFG4 & 4.16e-2(3.01e-3)-    & 1.14e-1(4.65e-3)-	& 2.91e-2(2.47e-3)-	& \textbf{2.90e-2(2.14e-3)}      \\ 
WFG5 & \textbf{7.12e-2(2.25e-3)}+	& 1.62e-1(2.78e-3)-	 & 7.27e-2(3.17e-3)-	& 7.21e-2(2.18e-3)    \\ 
WFG6 & \textbf{8.94e-2(5.12e-3)}+	&1.49e-1(4.89e-3)+	 & 8.34e-1(1.58e-3)- 	& 1.50e-1(3.20e-3)       \\ 
WFG7 & 3.02e-2(3.61e-3)-	& 1.78e-1(2.85e-3)-	& \textbf{2.08e-2(2.02e-3)}+	& 2.09e-2(1.96e-3)      \\ 
WFG8 & 1.14e-1(2.45e-3)-	& 2.03e-1(2.45e-3)-	& 1.08e-1(2.36e-3)-	& \textbf{1.05e-1(1.68e-3)}     \\ 
WFG9 & 5.11e-2(1.24e-3)-	& 1.45e-1(1.68e-3)-	& 3.56e-2(1.58e-3)-	& \textbf{3.55e-2(2.11e-3)}     \\ 
LSMOP5  &    9.36e+0(1.25e-1)-	&	9.25e+0(1.48e-1)-		&3.09e+0(4.65e-1)-		&  \textbf{1.00e+0(6.13e+0)}       \\ 
LSMOP6  &   8.10e-1(1.84e-2)-	&	8.08e-1(1.14e-2)-		& 8.12e-1(1.28e-2)-		& \textbf{5.16e-1(1.94e-2)}      \\ 
LSMOP7  &   1.21e+2(3.14e+1)-	&	1.33e+2(2.66e+1)-		& 1.73e+2(2.45e+1)-		& \textbf{6.01e-1(3.15e-2)}       \\ 
LSMOP8  &   6.10e-0(3.28e-2)-	&	6.34e-0(1.14e-2)-		& 5.77e-0(1.47e-2)-		& \textbf{2.09e-1(3.16e-3)}     \\ \hline
\multicolumn{1}{l}{+/-/=}   & 2/11/0        & 1/12/0       & 1/12/0         &            \\ \bottomrule
\end{tabular}
\caption{A comparison of results in terms of IGD on WFG and LSMOP instances}
 \label{tab:1} 
\end{table}
 
Note that in each of the optimization runs in this set of experiments, each MOEA performs 20,000 (the population size multiplied by the number of generations) evaluations of the objective functions, while EmoDM needs only 800, considerably reducing the computational or experimental cost in solving a new MOP.   

\subsection{Performance on Large-Scale Problems}
In the second set of the experiments, we include a state-of-the-art MOEA dedicated to large-scale MOPs, called ANSGA-II \cite{liu2022learning} for comparison on LSMOP instances with a search dimension ranging from 500 to 5000. In addition to EmoDM, we also consider three variants of EmoDM, namely EmoDM/A, EmoDM/M, and EmoDM/T, to examine the effect of the attention mechanism in EmoDM and the influence of the training data. The EmoDM/A is a variant of EmoDM without the mutual entropy-based attention mechanism. EmoDM/M is the EmoDM trained using evolutionary prompts equally contributed by three MOEAs, i.e., NSGA-II, MOEA/D, and SMS-EMOA. Finally, EmoDM/T is the EmoDM trained by the original training data plus the prompts from the test instances, i.e., those generated when NSGA-II optimizes instances LSMOPs 1-8 when it is tested on instances LSMOPs 5-8.

Table \ref{tab:3} presents the best IGD results obtained by four MOEAs, EmoDM and its three variants on LSMOP5-LSMOP8 instances of different search dimensions. As observed from Table \ref{tab:3}, EmoDM obtains the best IGD values in 13 out of 16 cases compared with NSGA-II, MOEA/D, SMS-EMOA and is only outperformed by ANSGA-II on three instances. Thus we can conclude that EmoDM remains effective in addressing MOPs when the search dimension increases up to 5000. 
Compared with its three variants, EmoDM achieves the best results in eight out of the 14 cases. In comparison, EmoDM/A obtains two best result, EmoDM/M obtains five best results, and EmoDM/T obtains eight best results. The fact that EmoDM/T slightly outperforms EmoDM appears reasonable, since EmoDM/T has seen the test problems in training. Note, however, that EmoDM/M is not able to outperform EmoDM even though its evolutionary prompts involve search processes from three MOEAs. This may imply that simultaneously learning search dynamics of different MOEAs is not beneficial. Interestingly, EmoDM achieves the best IGD values on four instances when the dimension increases to 5000.

\begin{figure}
  \centering
   \includegraphics[width=1.0 \linewidth]{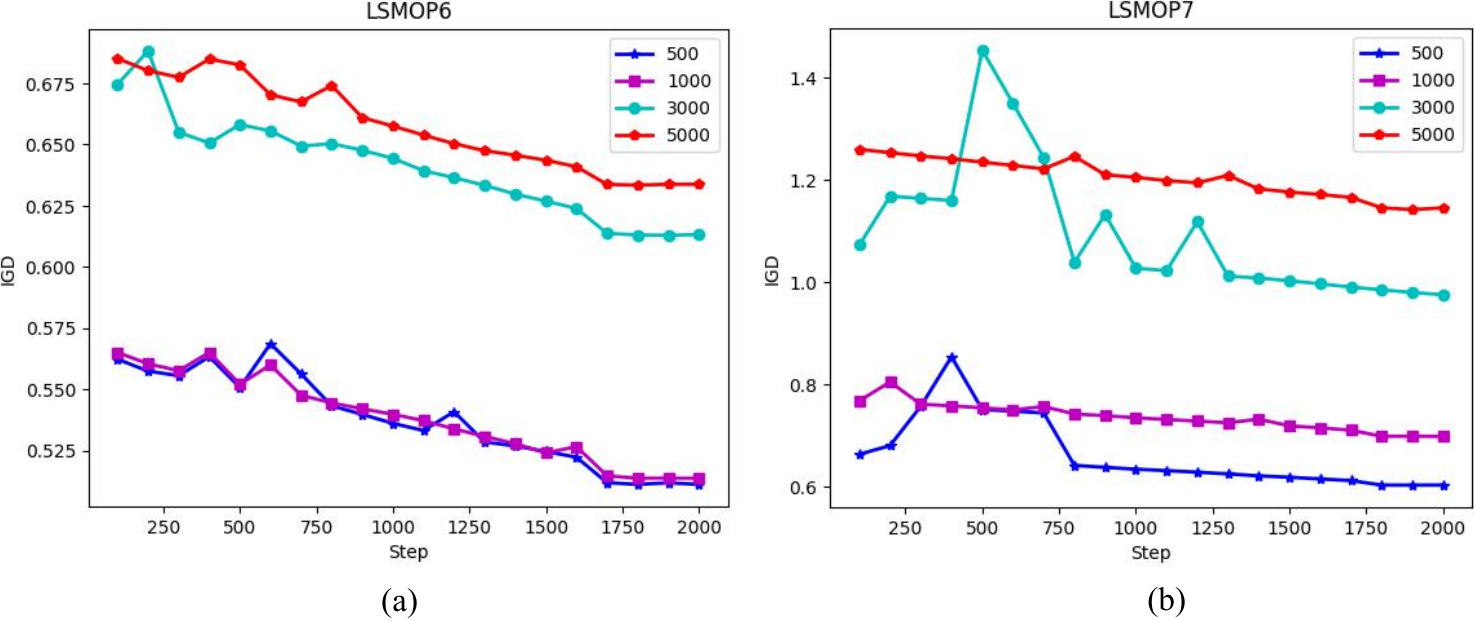}
   \caption{IGD profiles of EmoDM for Pareto set generation on LSMOP6 and LSMOP7 with 500, 1000, 3000, and 5000 decision variables.}
   \label{fig5}
 \end{figure}

Figure \ref{fig5} depicts the IGD profiles of EmoDM for Pareto set generation on LSMOP6 and LSMOP7 with 500, 1000, 3000, and 5000 decision variables, illustrating the search process of EmoDM. As shown in Fig. \ref{fig5}, although there are fluctuations during the early steps, the learning curve of EmoDM generally converges. For example, EmoDM experiences some fluctuations on LSMOP7 in the early reverse diffusion stage. Nevertheless, it manages to converge in the later stage. This can mainly be attributed to the adoption of a mutual information entropy-based sampling strategy during the inverse diffusion process in EmoDM, which allows it to adapt its search behaviors to LSMOP7.

\begin{table*}[]\tiny
\centering
\setlength{\tabcolsep}{8pt}
\begin{tabular}{lll|lllllllll}  \toprule
Problem    &  m           & d    & NSGA-II & MOEA/D & SMS-EMOA & ANSGA-II & EmoDM/A & EmoDM/M & EmoDM/T & EmoDM \\ \hline
\multirow{4}{*}{LSMOP5} &3& 500  & 9.83e+0	&3.13e+0	&9.66e+0	&1.73e+1   &   \textbf{1.07e+0} 	&1.12e+0	&1.08e+0	&1.08e+0      \\ 
                        &3& 1000 & 1.81e+1	&4.07e+0	&1.34e+1	&5.26e-1   &    3.72e-1	&2.93e-1	& \textbf{2.79e-1}	& \textbf{2.79e-1}      \\ 
                        &3& 3000 & 1.97e+1	&3.79e+0	&1.72e+1	&5.41e-1   &    4.52e-1	&\textbf{4.51e-1}	&4.62e-1	&4.59e-1     \\ 
                        &3& 5000 & 2.03e+1	&4.33e+0	&1.82e+1	&5.42e-1   &    4.95e-1	&4.88e-1	&\textbf{4.87e-1}	&\textbf{4.87e-1}         \\ \hline
\multirow{4}{*}{LSMOP6} &3& 500  & 8.10e-1	&8.15e-1	&8.09e-1	&\textbf{5.45e-1}   &    5.64e-1	&5.53e-1	&5.52e-1	&\text{5.51e-1}       \\ 
                        &3& 1000 & 1.88e+4	&7.18e+2	&4.89e+3	& 1.33e+0  &    1.25e+1	&0.23e+0	&\textbf{6.23e-1}	&\textbf{6.23e-1}   \\ 
                        &3& 3000 & 4.30e+4	&9.42e+2	&1.08e+4	&1.35e+0   &    1.13e+1	&8.14e-1	&\textbf{8.13e-1}	&8.14e-1        \\ 
                        &3& 5000 & 5.08e+4	&9.29e+2	&1.28e+4	&1.28e+0   &    1.48e+0	&\textbf{9.41e-1}	&9.57e-1	&\textbf{9.41e-1}       \\ \hline
\multirow{4}{*}{LSMOP7} &3& 500  & 1.23e+2	&1.92e+0	&1.45e+1	&6.12e-1   &    6.58e-1	&\textbf{6.02e-1}	&\textbf{6.02e-1}	&6.09e-1         \\ 
                        &3& 1000 & 2.78e+2	&1.04e+1	&2.04e+1	&\textbf{8.46e-1}  &  1.54e+0	&1.89e+0	&9.89e-1	&1.88e+0     \\ 
                        &3& 3000 & 6.46e+1	&9.84e+0	&2.04e+1	&\textbf{8.50e-1}& 2.77e+0	&2.24e+0	&2.02e+0	&2.02e+0        \\ 
                        &3& 5000 & 8.01e+1	&9.47e-1	&2.52e+1	&3.42e+1         & 4.98e-1	&\textbf{3.56e-1}	&\textbf{3.56e-1}	&\textbf{3.56e-1}        \\ \hline
\multirow{4}{*}{LSMOP8} &3& 500  & 6.21e-0	&5.82e-1	&6.51e-1	&3.13e-1         & \textbf{2.17e-1}	&2.39e-1	&\textbf{2.17e-1}	&\textbf{2.17e-1}     \\ 
                        &3& 1000 & 8.86e-1	&5.19e-1	&9.23e-1	&2.45e-1         & 2.67e-1	&\textbf{2.19e-1}	&2.21e-1	&2.20e-1   \\ 
                        &3& 3000 & 8.76e-1	&5.17e-1	&9.22e-1	&2.36e-1         & 5.26e-1	&2.35e-1	&2.23e-1	& \textbf{2.22e-1}  \\ 
                        &3& 5000 & 8.32e-1	&5.16e-1	&9.13e-1	&2.14e-1         & 2.94e-1	&2.64e-1	&\textbf{2.12e-1}	& \textbf{2.12e-1} \\ \hline
+/-/=                   & &      &    0/16/0     &  0/16/0      &   0/16/0     & 3/13/0 & 3/12/1      &  3/10/3       & 3/5/8        \\ \bottomrule
\end{tabular}
\caption{Best IGD results obtained by the compared algorithms and three other versions of EmoDM on LSMOP5-LSMOP8 instances}
\label{tab:3}
\end{table*}

In this set of the experiments, each MOEA requires 200,000 evaluations of the objectives, while EmoDM requires only 1,100 objective evaluations.  

\subsection{Further Analysis}
\label{sec:further}
Here we examine at first the performance of EvoDM when the dimension of the decision space of a new problem is lower than that of the problems for training. Note that it is impossible for EmoDM to solve a new problem having a decision space larger than that of the trained problems. Table \ref{tab:D} lists two sets of comparative experiments on Pareto set generation, where EmoDM is trained on 500-D, 1000-D, 3000-D, and 5000-D LSMOP instances, respectively, and then tested on a lower-dimensional instance. For example, "1000 $\to$ 500" means EmoDM is trained on 1000-D instances and then used to generate solutions for a new 500-D instance.

\begin{table}[]\tiny
\centering
\setlength{\tabcolsep}{10pt}
\begin{tabular}{lll|ll}
\toprule
Problem & 500$\to$500  & 1000$\to$500          & 3000$\to$3000 & 5000$\to$3000         \\ \hline
LSMOP5  & 1.121e+1 & \textbf{1.801e+0} & 1.250e+0  & \textbf{4.596e-1} \\ 
LSMOP6  & 2.105e+0 & \textbf{0.544e+0} & 9.130e-1  & \textbf{8.162e-1} \\ 
LSMOP7  & \textbf{6.173e-1} & 5.789e+0          & 3.748e+0  & \textbf{3.739e+0} \\ 
LSMOP8  & 3.298e-1 & \textbf{2.173e-1} & 3.293e-1  & \textbf{3.275e-1} \\ \bottomrule
\end{tabular}
\caption{Performance in terms of the average IGD value for Pareto set generation on three-objective LSMOP5-LSMOP8 instances}
\label{tab:D}
\end{table}

Next, we perform a sensitivity analysis of the hyperparameter $\xi$, which determines the number of function evaluations required by EmoDM. Table \ref{tab:F} presents the performance of EmoDM in terms of IGD on four 5000-dimensional three-objective LSMOP instances when $\xi$ is set to 1 (i.e., similarity checking is employed at every step), $T/10$, $ \left \lfloor log_{2}T \right \rfloor$ (the default value used in the experimental studies), and $T$ (checking in the first step only). From these results, we find that it does not improve its performance when EmoDM performs similarity checking at every step, implying that EmoDM can inherently save function evaluations without deteriorating its performance. 

\begin{table}[]\tiny
\centering
\setlength{\tabcolsep}{10pt}
\begin{tabular}{lllll}
\toprule
Problem & $\xi =1 $ & $\xi=T/10 $          & $\xi= \left \lfloor log_{2}T \right \rfloor  $& $\xi=T $        \\ \hline
LSMOP5  & 1.432e+1 & 5.321e-1  & \textbf{4.873e-1}  & 4.921e-0 \\ 
LSMOP6  & 5.325e+0 & 9.412e-1 & \textbf{9.410e-1}  & 1.432e+0 \\ 
LSMOP7  & 2.327e+1 & 1.324e+0 & \textbf{3.563e-1}  & 3.642e-1 \\ 
LSMOP8  & 3.480e-1 & 2.142e-1 & \textbf{2.128e-1}  & 2.329e-1 \\ \bottomrule
\end{tabular}
\caption{Performance in terms of the best IGD value on three-objective LSMOP5-LSMOP8 instances with 5000 dimensions}
\label{tab:F}
\end{table}

\section{Conclusion}
In this study, we propose a diffusion model for evolutionary multi-objective optimization. The basic idea is to use a diffusion model to learn evolutionary search behaviors with noise distributions in the forward diffusion and then solve now problems with reverse diffusion.  Additionally, mutual entropy-based attention is introduced to capture the decision variables that are most important for the objectives, making EmoDM more scalable to higher-dimensional problems. Comprehensive experiments demonstrate EmoDM can solve various MOPs of different dimensions up to 5000 with comparable or even better quality than popular MOEAs with a much smaller number of required function evaluations, indicating that the pre-trained EmoDM is a competitive and more efficient MOP solver compared to evolutionary algorithms. 

To the best of our knowledge, this is the first attempt that a diffusion model is trained for multi-objective evolutionary optimization. In the future, we will further investigate if a diffusion model can also learn other search behaviors such as reinforcement learning, or integrate search behaviors of multiple solvers. It is also of interest to know if this model can be extended to combinatorial optimization problems.
\appendix

\section*{Ethical Statement}

There are no ethical issues.

\section*{Acknowledgments}
We are grateful to Bo Yin, who helped conduct some of the experiments presented in this paper. 

\bibliographystyle{named}
\bibliography{ijcai23}

\end{document}